\documentclass[a4paper,fleqn]{cas-sc}

\usepackage[numbers]{natbib}
\usepackage{algorithm}
\usepackage{algpseudocode}
\usepackage{blindtext}
\usepackage{amssymb}
\usepackage{balance}
\usepackage{amsmath}
\usepackage{booktabs}
\usepackage{soul}
\usepackage[T1]{fontenc}
\usepackage{float}
\usepackage{hyperref}
\usepackage{enumitem}
\usepackage{tikzsymbols}

\def\tsc#1{\csdef{#1}{\textsc{\lowercase{#1}}\xspace}}
\tsc{WGM}
\tsc{QE}
\tsc{EP}
\tsc{PMS}
\tsc{BEC}
\tsc{DE}

\begin{document}
\let\WriteBookmarks\relax
\def\floatpagepagefraction{1}
\def\textpagefraction{.001}

\shorttitle{}

\shortauthors{Zia et~al.}

\title [mode = title]{A Context-aware Attention and Graph Neural Network-based Multimodal Framework for  Misogyny Detection}                      



%
\author[1]{Mohammad Zia Ur Rehman}[type=editor,
                        auid=000,bioid=1,
                        prefix=,
                        role=,
                        orcid=0000-0001-6374-8102]



\ead{phd2101201005@iiti.ac.in}


\credit{Conceptualization, Methodology, Software, Investigation,
Writing - Original Draft, Writing - review \& editing}

\affiliation[1]{organization={Department of Computer Science and Engineering},
    addressline={Indian Institute of Technology Indore}, 
    country={India}}

\author[2]{Sufyaan Zahoor}
\ead{sufyaan_2020bcse005@nitsri.net}
\credit{Software, Formal analysis, Investigation, Writing - Original Draft}

\author[2]{Areeb Manzoor}
\ead{areeb_2020bcse004@nitsri.net}
\credit{Software, Formal analysis, Investigation, Visualization, Writing - Original Draft}

\author[2]{Musharaf Maqbool}
\ead{musharaf_2020bcse003@nitsri.net}
\credit{Software, Formal analysis, Data Curation, Investigation, Writing - Original Draft}

\author[1]{Nagendra Kumar}[type=editor,
                        auid=000,bioid=1,
                        prefix=,
                        role=,
                        orcid=]
\ead{nagendra@iiti.ac.in}
\cormark[1]

\credit{Conceptualization, Methodology, Supervision, Writing - review \& editing}

\affiliation[2]{organization={Department of Computer Science and Engineering, National Institute of Technology Srinagar},
    country={India}}



\cortext[cor1]{Corresponding author}



\begin{abstract}
A substantial portion of offensive content on social media is directed towards women. Since the approaches for general offensive content detection face a challenge in detecting misogynistic content, it requires solutions tailored to address offensive content against women. To this end, we propose a novel multimodal framework for the detection of misogynistic and sexist content. The framework comprises three modules: the Multimodal Attention module (MANM), the Graph-based Feature Reconstruction Module (GFRM), and the Content-specific Features Learning Module (CFLM). The MANM employs adaptive gating-based multimodal context-aware attention, enabling the model to focus on relevant visual and textual information and generating contextually relevant features. The GFRM module utilizes graphs to refine features within individual modalities, while the CFLM focuses on learning text and image-specific features such as toxicity features and caption features. Additionally, we curate a set of misogynous lexicons to compute the misogyny-specific lexicon score from the text. We apply test-time augmentation in feature space to better generalize the predictions on diverse inputs. The performance of the proposed approach has been evaluated on two multimodal datasets, MAMI and MMHS150K, with 11,000 and 13,494 samples, respectively. The proposed method demonstrates an average improvement of 10.17\% and 8.88\% in macro-F1 over existing methods on the MAMI and MMHS150K datasets, respectively.
\end{abstract}




\maketitle

\section{Introduction}

\label{sec:sa_int}
Social media has become an indispensable aspect of our daily existence, offering a platform for open communication and idea-sharing. Despite its positive uses, the prevalence of offensive content is a notable concern~\cite{cohen2023enhancing,firmino2024improving}. Content is considered offensive if it includes any form of unacceptable language such as insults, threats, or bad language. Such offensive content is frequently directed at individuals or particular societal groups~\cite{ghosh2022sehc}, and there is a noticeable pattern where a substantial portion is aimed at women \cite{pamungkas2020misogyny}. 
Social media content that displays hatred, contempt, or prejudice against women is referred to as misogynistic content, and content that discriminates against women based on their gender is identified as sexist~\cite{parikh2021categorizing}. Such content can manifest in various forms including texts, images, and videos. Since women form a large social media user base, it is crucial to identify misogynistic content on social media as it helps in addressing online gender-based harassment and fostering a safer and more inclusive digital environment.

\subsection{Women on Social Media: A Statistical View}
The increasing number of users on social media platforms signifies the growing influence and widespread adoption of digital communication channels in today's interconnected world~\cite{putra2024semi}. 
As of October 2023, social media has infiltrated nearly 61.4\% of the global population with nearly 4.95 billion users. Out of all internet users worldwide, 93.5\% use social media. 
There has been a 4.5\% increase in social media users within a single year, accounting for almost 215 million new users~\cite{smartinsightsGlobalSocial}. The gender distribution on popular social media platforms like Snapchat, Instagram, and Facebook varies slightly, with Snapchat having the highest percentage of female users at 51.00\%, Instagram has around 48.20\% female users and Facebook has 43.70\%~\cite{statistaSocialPlatforms}.
These statistics indicate that women actively participate in online communication on social media; therefore, it is imperative to find solutions to safeguard women from online harassment by filtering out misogynistic and sexist content.

\subsection{Existing Approaches and Challenges} 
Examining and tackling the growing prevalence of online misogyny and sexism is crucial. Multiple potential solutions exist for identifying misogyny and sexism in the literature. 
Early approaches to misogyny detection relied on unimodal feature engineering such as N-grams~\cite{anzovino2018automatic},  Bag of Words~\cite{pamungkas2018automatic}, Term Frequency-Inverse Document Frequency~\cite{bakarov2018vector}, and the utilization of word embeddings~\cite{garcia2021detecting} for text analysis. Existing works have also explored linguistic and syntactic handcrafted features such as part of speech, sentence length, sentiment, and hate lexicons to flag misogyny~\cite{anzovino2018automatic, garcia2021detecting}. Although these methods were foundational, recent advancements have shifted towards more sophisticated techniques, such as transformer-based methods~\cite{samghabadi2020aggression}, multimodal analysis~\cite{rizzi2023recognizing}, and larger and diverse datasets~\cite{fersini2022semeval}, allowing for more nuanced and context-aware misogyny detection across a broader spectrum of content.

\begin{figure}
    \centering
    \includegraphics[width=7.15cm, height=8.5cm]{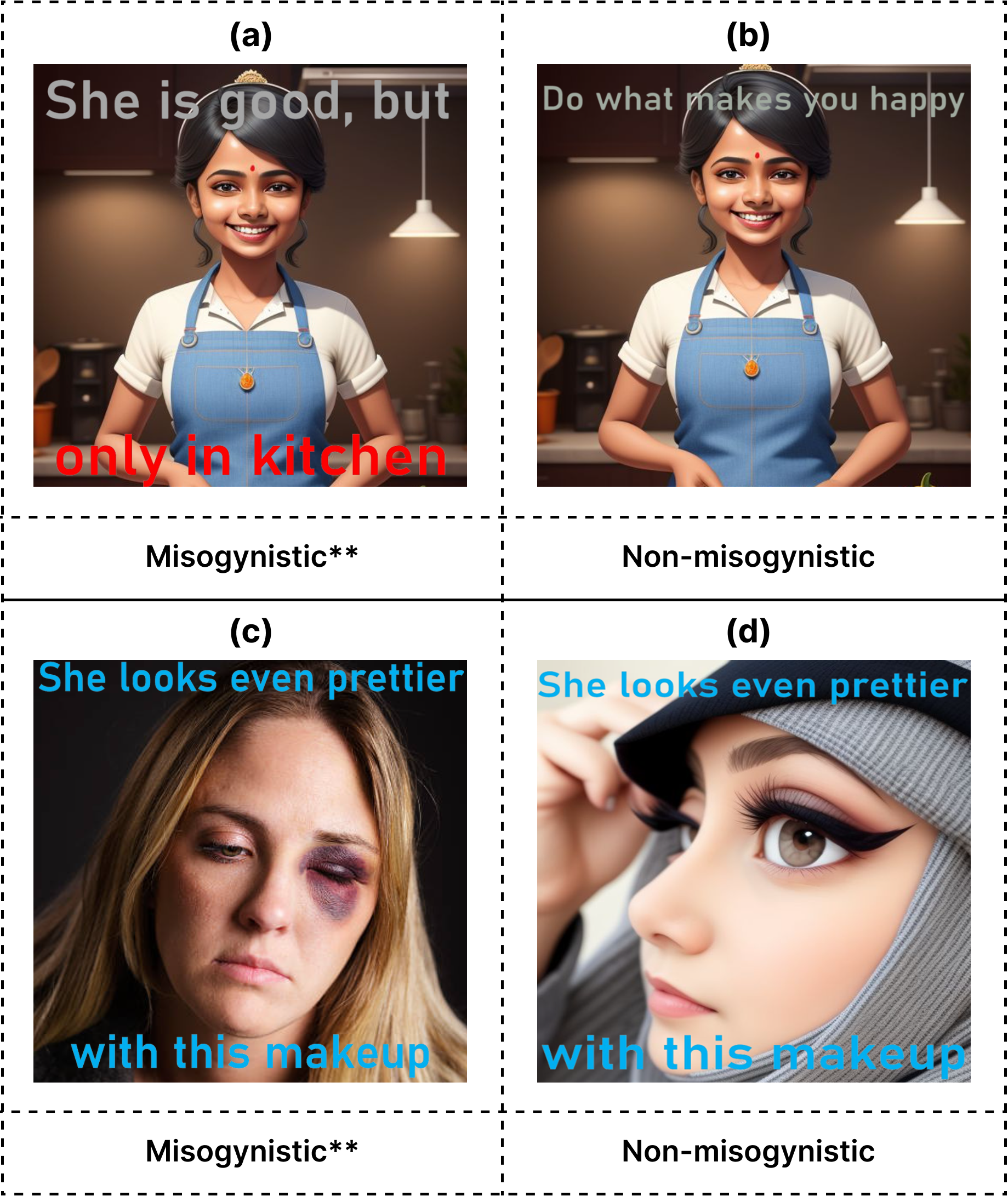}
    \caption{Examples of misogynistic and non-misogynistic content}
    \label{fig:examples}
\end{figure}

A significant volume of misogynistic content is disseminated through multimodal memes that combine images and text. This presents a challenge as
images that might appear misogynistic in one context may not be interpreted as such in another. \autoref{fig:examples} illustrates an instance where the interpretation of an image varies when combined with different contexts. Figure 1(a) and Figure 1(b) have the same image but different texts, giving them different class labels. Likewise, identical text can have different labels when paired with different images. Figure 1(c) and Figure 1(d) have identical text but with different images which changes the class label in both the memes. Misogynistic content can also be disguised in sarcasm or irony, making detection a complex task. Therefore, for the aforementioned reasons, comprehending both modalities through context-aware cross-modal interaction while also concentrating on discriminative information within individual modalities is crucial in misogyny detection.   Additionally, it can adapt to new slang or lexicons, further complicating identification. Our proposed approach revolves around tackling these challenges.\renewcommand{\thefootnote}{$ $} \footnote{**Offensive content warning: The presented image is just for illustration. We do not promote the views presented in the image.}

\subsection{Research Objectives}
As mentioned in the previous section, numerous challenges are present in the domain of multimodal content analysis, underscoring the complex task of effective integration of diverse modalities such as text and images. Leveraging our prior discourse on these challenges, the primary objectives of this research are as follows:
\begin{itemize}
    \item{To explore a novel multimodal context-aware cross-attention mechanism for offensive content detection against women.} 
    \item{Exploring the influence of feature space test-time augmentation on the generalizability of predictions for multimodal content.}
    \item{Modeling a unified framework that learns different features through dedicated modules and exploring its effectiveness for multimodal misogyny detection.}
   
\end{itemize}
\subsection{A Glimpse into the Proposed Solution Framework}
Reflecting on the aforementioned objectives and challenges, we propose a multimodal context-aware attention-based approach for misogyny and sexism detection. The proposed approach employs a multi-faceted strategy where a graph neural network is used for unimodal feature reconstruction to focus on discriminative information in each modality, and a context-aware attention module provides multimodal features. Our approach also leverages a comprehensive set of additional features that include content-specific text and image features. Text features include misogyny-specific lexicon score (MSL) and toxicity features, whereas image features include image caption embeddings and NSFW features. We employ test time augmentation (TTA) to help the model generalize efficiently to variations in the test data that may not have been present in the training data. To our knowledge, none of the existing studies harness the collective advantages of these techniques for detecting offensive content targeting women.
In essence, our approach capitalizes on the synergy of these features that help improve the model's performance by deriving valuable insights from multimodal content.

\subsection{Key Contributions}
The primary contributions of this research study are as follows:
\begin{enumerate}
    \item{Our proposed multimodal misogyny detection approach explores the fusion of features through dedicated modules where multimodal features are learned through a novel multimodal context-aware cross-attention module, unimodal features are refined through a graph-based module, and another module learns the content-specific features.}
    \item{The proposed context-aware cross-attention effectively emphasizes the interaction between crucial regions of an image and words of the text through the integration of the global context of the image.}
    \item{We employ an additive random perturbations-based test time augmentation method to help enhance its ability to generalize to variations in the test data. It is achieved through the application of diverse augmentations to the test samples, the model is subjected to a broader range of inputs, enhancing its robustness. } 
    \item{We perform extensive analysis of the proposed framework on two diverse multimodal datasets. Experimental evaluations demonstrate and affirm the efficacy of our approach.}
\end{enumerate}

\section{Related Work}
\label{sec:sa_rw}
Scholarly investigations have primarily focused on two approaches for misogyny and sexism detection: text-based and multimodal methods. The text-based approach involves scrutinizing linguistic content to identify offensive language.
  Conversely, the multimodal approach incorporates diverse modalities such as images and their textual annotations. This approach acknowledges the multifaceted nature of online communication, recognizing that misogynistic content can manifest not only in words but also in visual elements. In this section, we present existing works related to unimodal and multimodal approaches.

\subsection{\textit{Unimodal Misogyny and Sexism Detection}}
Guo et al.~\cite{guo2023coco} employ a transformer method to detect sexism in textual content. BERT+BiLSTM is used for local and global features, and focal loss is used to improve the performance further. Data augmentation is additionally employed to augment the sample count. Due to their ability to provide contextual embeddings, transformer-based techniques have been extensively investigated for misogyny detection in the existing works. Finetuning of transformer-based models for downstream tasks has shown performance improvement~\cite{attanasio2022benchmarking,calderon2023enhancing,muti2022misogyny}. Muti et al.~\cite{muti2022checkpoint} provide an approach for misogyny detection in text-based content for English, Spanish, and Italian languages. They train ten transformer-based models for different combinations of languages. Dehingia et al.~\cite{dehingia2023violence} compare the BERT model with 
 Machine Learning (ML) methods such as Naive Bayes, Logistic Regression (LR), and Support Vector Machine (SVM). They utilize the Term Frequency - Inverse Document Frequency (TF-IDF) approach for ML methods. Their observation indicates that BERT performs better than ML methods. Unimodal approaches, particularly those relying solely on text, have a limited ability to understand the broader context surrounding a statement. As online content becomes increasingly multimedia-rich, unimodal approaches struggle to analyze images, videos, or a combination of different modalities effectively. Offensive content can be shared through visual elements or a combination of text and images, requiring a broader approach.

\subsection{\textit{Multimodal Misogyny Detection}}
Zhang \& Wang~\cite{zhang2022srcb} explore the relationship between vision and language by studying the effectiveness of multimodal pretrained models. Findings from the study indicate that multimodal fine-tuning generally performs better than unimodal approaches for both UNITER and CLIP, with CLIP excelling due to extensive pretraining. XGBoost classifier competes with fine-tuning, particularly by leveraging CLIP's image features. 
Muti et al.~\cite{muti2022unibo} address a task focused on detecting misogyny in memes using both text and image data. The proposed MMBT model, which integrates text and image embeddings, employed BERT for text and CLIP for image encoding. This model demonstrates superior performance compared to models utilizing only one modality. 
Gu et al.~\cite{gu2022mmvae} introduce an approach to detect and classify misogyny in multimedia content using both text and images. The proposed multimodal-multitask variable autoencoder (MMVAE) model combines an image-text embedding module, variable autoencoder module, and multitask learning module. Pretrained models like BERT, ResNet-50, and CLIP are used for embedding, and the VAE combines them into a latent representation. The MMVAE outperforms single-modal baselines such as BERT and ResNet-50. 

Gomez et al.~\cite{gomez2020exploring} present an innovative contribution to hate speech detection by introducing the MMHS150K dataset, containing 1,50,000 Twitter image-text pairs annotated as hate speech or not. One of the labels in the dataset is sexism. The study introduces the Feature Concatenation Model (FCM), Spatial Concatenation Model (SCM), and Textual Kernels Model (TKM). 
A few other existing works focus on different approaches for multimodal hate speech detection such as transformer-based ~\cite{yuan2024rethinking}, ensemble learning~\cite{mahajan2024ensmulhatecyb}, and pretrained models~\cite{bhandari2023crisishatemm}. 

Most of the aforementioned approaches have shown improvement over unimodal approaches except the approach presented by Gomez et al.~\cite{gomez2020exploring}. However, these approaches utilize text-embeddings and visual features only while neglecting important features such as toxicity features and image captions. These features may provide useful insights into the content. Additionally, the generalization of predictions of test data is also one of the limitations that can be overcome by TTA.

\section{Methodology}
\label{sec:sa_meth}
\begin{figure*}
    \centering
    \includegraphics[width=16.5cm, height=9.1cm]{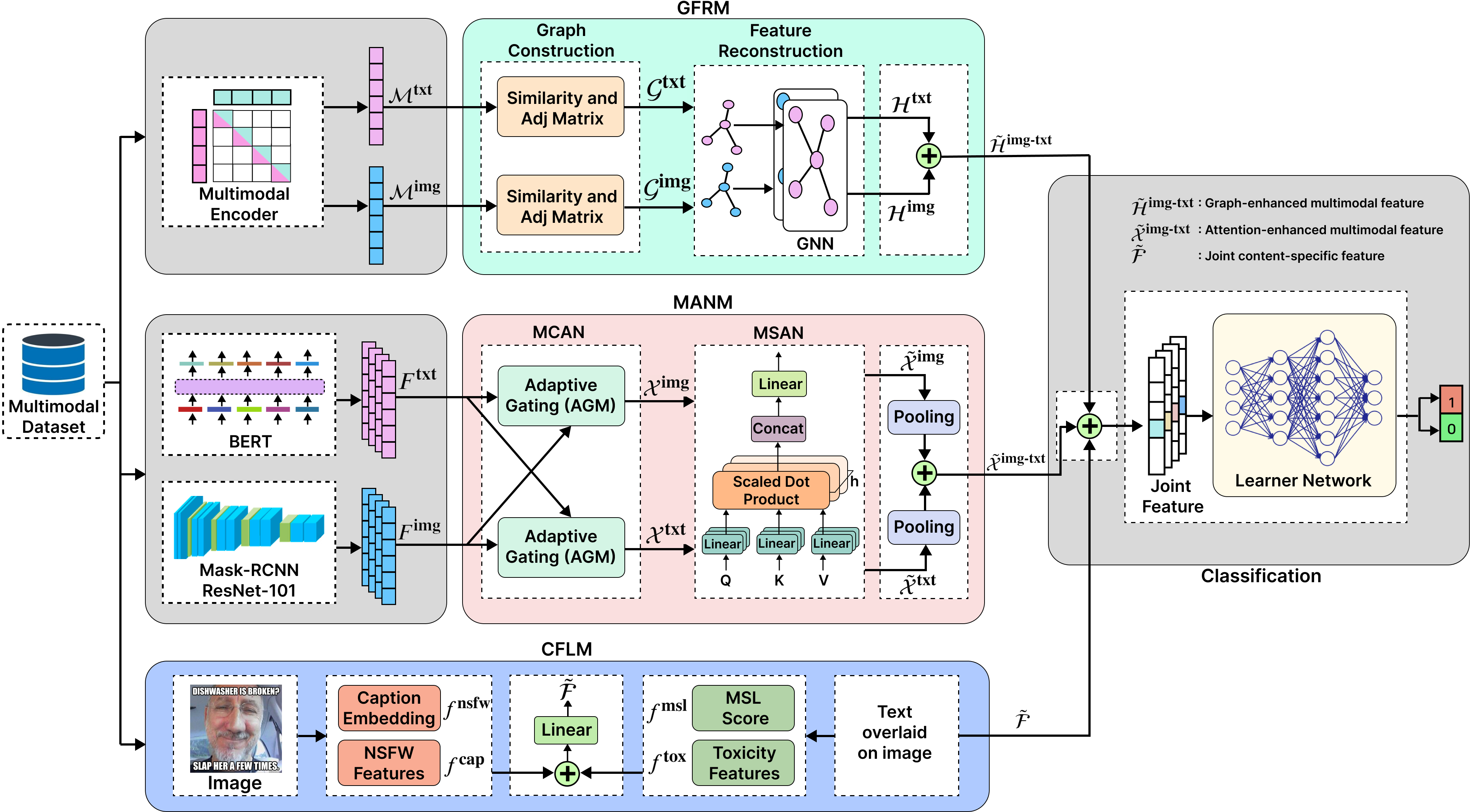}
    \caption{System architecture of the proposed method: Features for Graph-based Features Reconstruction Module (GFRM) and Multimodal Attention Module (MANM) are extracted in separate dedicated modules. Additionally, the Content-specific Feature Learning Module (CFLM) extracts and concatenates extra content-based features. }
    \label{fig:architecture}
\end{figure*}

This section presents a detailed description of the proposed approach for detecting misogynistic instances on social media. The architecture explaining the proposed method is shown in \autoref{fig:architecture}. We first apply data cleaning and extract features for Graph and Attention modules. These features are passed through corresponding networks, MANM and GFRM. Content-specific features such as toxicity features from text and captions from images are also extracted to gain insights into the content. Content-specific features are passed through CFLM. Next, the refined features are then combined to form a unified representation for final classification. Subsequent sections elaborate on the proposed methodology.

\subsection{Data Preprocessing}
Preprocessing of data involves the cleaning of both images and text. Image cleaning entails the removal of text from images through processes like image masking and inpainting. Text cleaning is conducted to improve the quality and standardize textual data. The preprocessed content aids in accurate feature extraction.

\subsubsection{Image Cleaning}
The multimodal datasets we use have text superimposed on the images. We employ image cleaning through image inpainting technique proposed by Yu et al.~\cite{yu2019free} to remove the text from images. The cleaning process includes masking and inpainting.  In the masking step, text regions are identified in the image and a binary image mask is created that isolates text from the inherent visual content. In the inpainting step, the removed areas are reconstructed to ensure that inpainted regions blend seamlessly with nearby regions, resulting in visually appealing and coherent images. This facilitates the generation of more accurate and relevant feature representations as the emphasis is directed towards the intrinsic visual content of the image.

\subsubsection{Text Cleaning}
The text cleaning process involves several steps to enhance the quality and consistency of textual data. It includes converting text to lowercase and removing unwanted characters, symbols, and excessive white spaces. The process extends to the removal of URLs or usernames. 

\subsection {\textit{Feature Extraction}}
To enhance the capabilities of our framework, we incorporate diverse feature extraction techniques for images and text alongside content-specific features, providing a more comprehensive understanding of the data.
\subsubsection{\textit{Feature Extraction for Graph-based Module}}
For the proposed GFRM, we extract image and text features through the Contrastive Language-Image Pre-training (CLIP) model~\cite{radford2021learning}. CLIP is a versatile model that learns representations by aligning images and their associated textual descriptions. CLIP encoder employs a transformer architecture similar to Vision Transformers (ViT) to extract visual features and utilizes a causal language model to capture text features. The encoder processes the image-text pair and subsequently maps image-text to a shared latent space with equivalent dimensions, i.e., image embeddings $E^{\text{img}} \in \mathbb{R}^{512}$ and text embeddings $E^{\text{txt}} \in \mathbb{R}^{512}$. 

\subsubsection{\textit{Feature Extraction for Attention Module}}
The proposed MANM requires region embeddings from images and word embeddings from the text. We utilize pretrained BERT~\cite{devlin2018bert}, a transformer-based model, to generate word embeddings. BERT is designed to capture the bidirectional context for each word. It takes tokens (words) as inputs and generates word embeddings of 768 dimensions.

\begin{equation}
    F^{\text{txt}} = \overrightarrow{\text{BERT}}(w)
\end{equation}
Here, $F^{\text{txt}}$ denotes the token feature matrix for a sentence, $F^{\text{txt}}=\{q_k\}_{k=1}^{100}$, where $q_k \in \mathbb{R}^{768}$ represents the token embedding of the $k$-th token, and $w$ denotes the tokens.
We consider a maximum of 100 tokens.

For visual feature extraction, MASK RCNN-based Detectron2 is employed that provides region-based features~\cite{wu2019detectron2}. This model utilizes ResNet-101 as its backbone~\cite{He_2016_CVPR}. ResNet-101-based MASK RCNN is a popular architecture for image feature extraction and instance segmentation.
For extracting visual embeddings, we need the features from various regions in the image. In this paper, we refer to this model as MASK RCNN.
The region proposal network detects object regions (bounding boxes).
The region bounding boxes are used to extract features. MASK RCNN ensures that the features are consistently sized and aligned with the regions.
\begin{equation}
\text{$F^{\text{img}}$} = \text{MASK\_RCNN(B)}
\end{equation}
The $F^{\text{img}}$ represents the features extracted from the region proposals $B$. These features capture the information about the object contained within the regions.
We thus get a feature matrix $F^{\text{img}} = \{r_k\}_{k=1}^{100}$ for each image, where $r_k \in \mathbb{R}^{1024}$ denotes the $k$th region's embedding of a given image.

\subsubsection{\textit{Feature Extraction for CFLM }}
\label{sec:cflm_feat}
We incorporate four additional features namely, toxicity features, image caption features, NSFW features, and misogyny-specific lexicon scores to enhance the text analysis capabilities. These features are learned in a content-specific feature learning module. 


\textit{Toxicity features}: We have used toxic-Bert to generate the toxicity features and their score for the text\renewcommand{\thefootnote}{$3 $}\footnote{\label{detoxify}\href{Github. https://github.com/unitaryai/detoxify}{Github. https://github.com/unitaryai/detoxify}}. These features include threat, insult, toxic, identity hate, obscene, and severe toxic. Toxicity features are represented by $f^{\text{tox}}$ where $f^{\text{tox}} \in \mathbb{R}^{6}$.

\textit{NSFW features}: NSFW model is used to identify indecent content\renewcommand{\thefootnote}{$4 $}\footnote{\label{NSFW}\href{Github. https://github.com/GantMan/nsfw\_model}{Github. https://github.com/GantMan/nsfw\_model}}. It uses an image and gives probabilities of five classes which are drawing, hentai, neutral, porn and sexy. We denote NSFW features with $f^{\text{nsfw}}$ where $f^{\text{nsfw}} \in \mathbb{R}^{5}$.

\textit{Misogyny-specific Lexicon score (MSL)}: Misogynous text might contain a few words that are explicitly offensive towards women. Through the MSL score, we aim to compute the number of such words present in a text. To achieve this,  we collect a set of misogynous lexicons~\cite{plaza2020detecting,bassignana2018hurtlex}. This set is further extended by adding synonyms, resulting in a large set of misogynous lexicons. To compute the msl score, we tokenize a given sentence and match each token to this list. After calculating the number of misogynous lexicons in each sentence, this number is normalized using min-mix normalization. We denote the msl score with $f^{\text{msl}}$.

\textit{Image Caption Features}: To leverage additional information from the image, we use image caption features. Image captions are generated using pretrained Bootstrapping Language-Image Pre-training (BLIP) method~\cite{li2022blip}. It is trained on the COCO dataset for image-captioning and it uses ViT-large as the backbone. The features for generated captions are extracted using the CLIP text feature extractor. Caption features are denoted with $f^{\text{cap}}$ where $f^{\text{cap}} \in \mathbb{R}^{512}$.

\subsection{Multimodal Attention Module (MANM)}
Attention mechanisms play a crucial role in Deep Learning, especially in Computer Vision and Natural Language Processing (NLP). 
We first apply a novel context-aware cross-attention to get cross-modal representations of both modalities and then self-attention is applied to focus on pertinent features within cross-modal features. Next, mean pooling is applied to represent the aggregate sequence embedding. 

\subsubsection{Multimodal Context-aware Attention (MCAN)}
\begin{figure*}
    \centering
    \includegraphics[width=14cm, height=4.46cm]{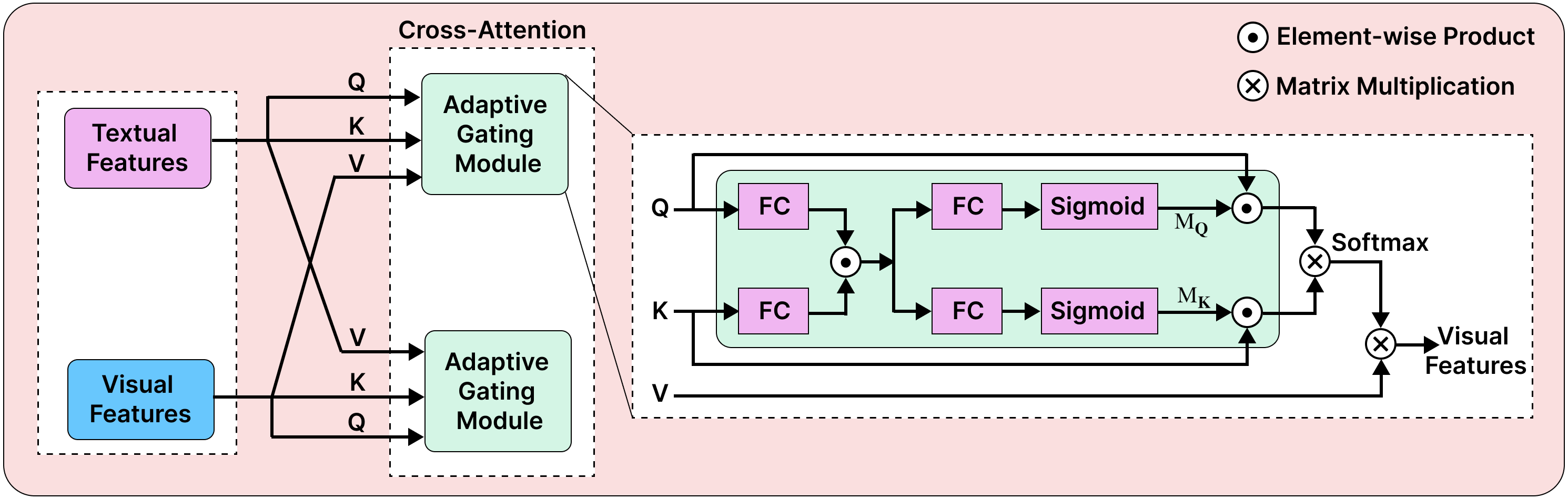}
    \caption{Adaptive gating-based cross attention for MCAN}
    \label{fig:mcan}
\end{figure*}

Multimodal Context-aware Attention (MCAN) enhances the representations of images and texts by exploiting the inter-modal context information. \autoref{fig:mcan} illustrates the MCAN mechanism. MCAN utilizes context-aware region representations of images by leveraging semantic and spatial relations among regions~\cite{qu2020context}. Internally, it uses cross-attention that aids in propagating the contextual clues between text and images~\cite{li2021selfdoc}. Before applying MCAN, the dimensions of image and text embeddings are aligned by projecting each image's region embeddings \(F^{\text{img}}\) into the same dimension as text embeddings, i.e., 768. The projection is done using the fully-connected dense layer as follows:
\begin{equation}
\label{eq:region_proj}
    F^{\text{img}}_{\text{proj}} = F^{\text{img}}\cdot W_i + b_i
\end{equation}

where, \(W_i \in \mathbb{R}^{1024 \times 768}\) is a learnable projection matrix and $b_i$ is a bias vector. Hence, we have image-embeddings $F^{\text{img}}_{\text{proj}}=\{r_k\}_{k=1}^{100}$ and text embeddings 
$F^{\text{txt}}=\{q_k\}_{k=1}^{100}$, where $r_k, q_k \in \mathbb{R}^{768}$.

Next, the query, key, and value is computed for the image as well as the text. Let us assume for an image-text pair the query, key, and  value vectors are defined as follows:
\begin{equation}
\label{eq:qkv_img}
\left\{
\begin{aligned}
Q_i = F^{\text{img}}_{\text{proj}} \cdot (W_{Q_i})\\
K_i = F^{\text{img}}_{\text{proj}} \cdot (W_{K_i})\\
V_i = F^{\text{img}}_{\text{proj}} \cdot (W_{V_i})
\end{aligned}
\right.
\end{equation}

\begin{equation}
\label{eq:qkv_text}
\left\{
\begin{aligned}
Q_t = F^{\text{text}} \cdot (W_{Q_t})\\
K_t = F^{\text{text}} \cdot (W_{K_t})\\
V_t = F^{\text{text}} \cdot (W_{V_t})
\end{aligned}
\right.
\end{equation}
where, ($Q_i$, $K_i$, $V_i$) are the query, key, and value of the image modality, ($Q_t$, $K_t$, $V_t$) represent the same the for the text, ($W_{Q_i}, W_{K_i}, W_{V_i}$) denote the weight matrices associated with the query, key, and value projections of image, ($W_{Q_t}, W_{K_t}, W_{V_t}$) are weight matrices associated with the query, key, and value projections of text, and ($F^{\text{img}}, F^{\text{text}}$) denote the context-aware region embeddings of image and word embeddings for text, respectively.\\

\hspace{-6mm}\textit{\textbf{3.3.1.1. Cross-Attention }}: We employ cross-attention between the two modalities by taking the value from the one while using the other for the query and key~\cite{li2021selfdoc}. The two outputs are provided by the cross-attention, $\mathcal{X}^{\text{img}}$ and $\mathcal{X}^{\text{txt}}$ that denote the cross-attended image features and cross-attended text features, respectively. 
The cross-attention operations to generate $\mathcal{X}^{\text{img}}$ and $\mathcal{X}^{\text{txt}}$ are defined as follows:

\begin{equation}
\label{eq:cross_img}
    \text{Cross\_Att}(Q_t,K_t,V_i) = \text{softmax}\left(\frac{Q_t.K_t^T}{\sqrt{d_k}}\right) \cdot (V_i)
\end{equation}

\begin{equation}
\label{eq:cross_text}
    \text{Cross\_Att}(Q_i,K_i,V_t) = \text{softmax}\left(\frac{Q_i.K_i^T}{\sqrt{d_k}}\right) \cdot (V_t)
\end{equation}

where T is the transpose, Cross\_Att($\cdot$) denotes cross-attention operation and \(d_k\)  denotes the key vector's dimension.\\

\hspace{-6mm}\textit{\textbf{3.3.1.2.  Gating Mechanism }}:
Instead of applying dot-product between a query $Q$ and key $K$ as shown in \autoref{eq:cross_img} and \autoref{eq:cross_text}, we use Adaptive Gating Module (AGM) with fusion strategy~\cite{qu2020context}. AGM aims to dynamically regulate the information exchange between queries (\( Q \)) and keys (\( K \)) with the goal of enhancing meaningful interactions and restraining noise or irrelevant information. Within AGM, queries and keys undergo initial projection into a shared space, followed by fusion. The fusion operation is expressed as follows:

\begin{equation}
    G = (Q \cdot W_{Q_G} + b_{Q_G}) \odot (K \cdot W_{K_G} + b_{K_G})
\end{equation}

where, $G$ is the fusion result, ($W_{Q_G}$, $W_{K_G}$) are learnable projection matrices, and ($b_{Q_G}$, $b_{K_G}$) are bias vectors,and $\odot$ is the element-wise product.

Gating masks, \(M_{Q}\) and \(M_{K}\), corresponding to \(Q\) and \(K\) are generated by applying fully-connected layers followed by the sigmoid function as shown in \autoref{eq:maskQ} and \autoref{eq:maskK}.
\begin{equation}
\label{eq:maskQ}
    M_{Q} = \sigma(G \cdot W_{Q_M} + b_{Q_M})
\end{equation}
\vspace{-6mm}
\begin{equation}
\label{eq:maskK}
    M_{K} = \sigma(G \cdot W_{K_M} + b_{K_M})
\end{equation}
Here, $\sigma(\cdot)$ denotes the sigmoid function, ($W_{Q_M}$, $W_{K_M}$) are learnable weights, and ($b_{Q_M}$, $b_{K_M}$) bias vectors.
The obtained gating masks, \(M_{Q}\) and \(M_{K}\), are used to control the information flow of the original queries and keys in the attention mechanism. The dot-product attention is applied with this controlled flow as shown in \autoref{eq:final}, hence providing context-enhanced cross-modal representation.
\begin{equation}
\label{eq:final}
    \mathcal{X} = \text{Cross\_Att}(M_{Q} \odot Q, M_{K} \odot K, V)
\end{equation}
Here, $\mathcal{X}$ is the result of the gated cross-attention. 

This mechanism allows the model to selectively focus on meaningful interactions, suppress irrelevant or noisy information, and provide a context-enhanced cross-attention mechanism for various tasks.\\


\hspace{-6mm}\textit{\textbf{3.3.1.3 Context-aware Image Region Features }}: Context-aware image region embeddings are calculated by performing the element-wise dot product between region embeddings and position embeddings~\cite{qu2020context}.
We encapsulate semantic embeddings with \(F^{\text{img}}_{\text{proj}} \in \mathbb{R}^{R \times D}\), and position embeddings with \({\hat{I}^{\text{pos}}} \in \mathbb{R}^{R \times D}\). $\hat{I}^{\text{pos}}$ is computed by taking absolute position encodings~\cite{wang2019position} of image regions, aspect ratio and area of image~\cite{qu2020context}. These values are normalized and we get a vector \(I \in \mathbb{R}^{6}\). To obtain absolute position embeddings $\hat{I}^{\text{pos}}$, $I$ is projected to the same dimension as region embeddings, i.e., 768 using a fully-connected dense layer as shown in \autoref{eq:pos_embed}.
\begin{equation}
\label{eq:pos_embed}
    \hat{I}^{\text{pos}} = \sigma(W_I \cdot I + b_I) 
\end{equation}

Here, \(W_I \in \mathbb{R}^{768 \times 6}\) is weight matrix and $b_I$ is bias.
Next, fusion is performed between image region embeddings and positional embeddings to obtain context-aware region features, $\tilde{F}^{\text{img}}$, as shown in \autoref{eq:context}.

\begin{equation}
\label{eq:context}
 \tilde{F}^{\text{img}} = F^{\text{img}}_{\text{proj}} \odot \hat{I}^{\text{pos}} 
\end{equation}

Hence, in \autoref{eq:qkv_img} we pass context-aware image features $\tilde{F}^{\text{img}}$ in place of $F^{\text{img}}_{\text{proj}}$ which return cross-modal image features $\mathcal{X}^{\text{img}}$. $\tilde{F}^{\text{img}}$ leverages the global spatial features as well as semantic region features which help in getting context-enhanced cross-modal features.

\subsubsection{Multihead Self-Attention (MSAN)}
Multihead Self-Attention (MSAN) enhances the self-attention mechanism by employing multiple sets of queries, keys, and values, which are referred to as attention heads~\cite{vaswani2017attention}. Each attention head processes the input independently, capturing different aspects and relationships within the sequence.
The outputs of the attention heads are then concatenated to create a comprehensive representation that integrates information from different attention subspaces. We apply MSAN on cross-modal features obtained through MCAN to focus on pertinent information within cross-modal features. In MSAN each attention head performs the self-attention operation as shown in \autoref{eq:self_msan}.

 \begin{equation}
 \label{eq:self_msan}
     Attention(Q_k,K_k,V_k) = \text{softmax}\left(\frac{Q_kK_k^T}{\sqrt{d}}\right)V_k
 \end{equation}
Here, $Q_k$, $K_k$, and $V_k$ denote the query, key, and value of $k$th head. Outputs from all the heads are concatenated and subjected to linearly transformation to get the final representation.
\begin{equation}
\text{concat\_output} = \text{concat}(\text{head}_1, \text{head}_2, \ldots, \text{head}_H)
\end{equation}

Here, $\text{head}_k$ denote the output from $k$-th head. In the MSAN, we pass cross-modal image and text features, $\mathcal{X}^{\text{img}}$ and $\mathcal{X}^{\text{txt}}$ which returns self-attended features, $\mathcal{\tilde{X}}^{\text{img}}$ and $\mathcal{\tilde{X}}^{\text{img}}$ as shown in \autoref{eq:msan}:

\begin{equation}
\label{eq:msan}
\mathcal{\tilde{X}}^{\text{img}}, \mathcal{\tilde{X}}^{\text{txt}} = \text{MSAN} (\mathcal{X}^{\text{img}}, \mathcal{X}^{\text{txt}})
\end{equation}

where, $\mathcal{\tilde{X}}^{\text{img}}, \mathcal{\tilde{X}}^{\text{txt}} \in \mathbb{R}^{128 \times 256}$. 

\subsubsection{Pooling and Concatenation}
Next, we apply max pooling on $\mathcal{\tilde{X}}^{\text{img}}$ and $\mathcal{\tilde{X}}^{\text{txt}}$ to get sequence vectors such that both vectors are of size $\mathbb{R}^{256}$. Subsequently, resultant vectors are concatenated to get the final multimodal attented feature vector of the image-text.

\begin{equation}
\mathcal{\tilde{X}}^{\text{img-txt}} = \text{concat}(\text{max\_pool}(\mathcal{\tilde{X}}^{\text{img}}, \mathcal{\tilde{X}}^{\text{txt}}))
\end{equation}

Here, $\mathcal{\tilde{X}}^{\text{img-txt}} \in \mathbb{R}^{512}$ denotes the final joint representation of multimodal attended image-text features.

\subsection{Graph-based Feature Reconstruction Module (GFRM)}
\begin{figure}[t!]
    \centering
    \includegraphics[width=8cm, height=7.4cm]{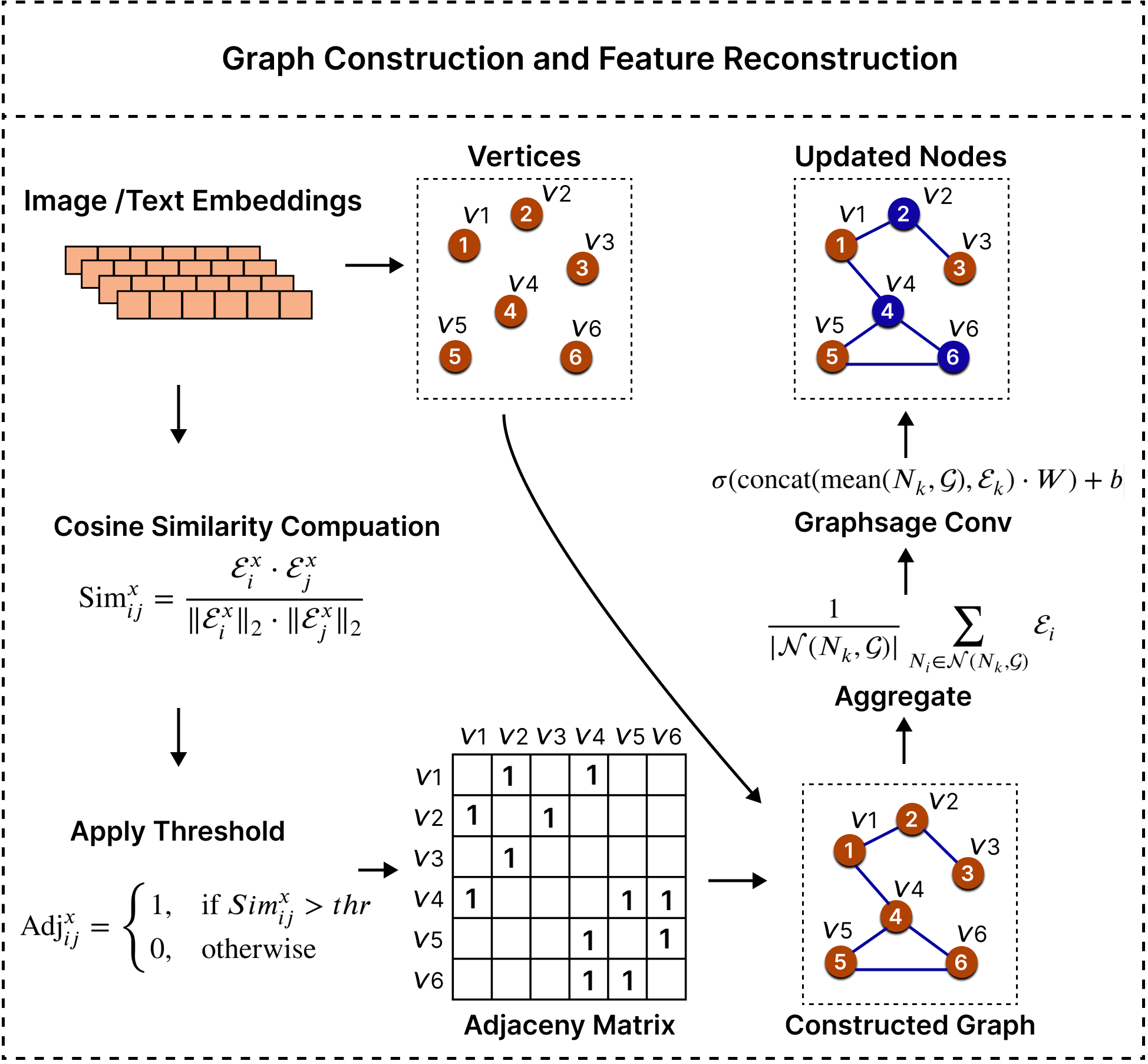}
    \caption{Graph construction and updating: The vertices are features, and an edge between two vertices is constructed based on cosine similarities of features.}
    \label{fig:graph}
\end{figure}

The Graph-Based Feature Reconstruction Module enhances unimodal features through a graph neural network. Feature reconstruction occurs within the context of node neighborhoods, which are determined by the similarity of each node to others. To achieve this, we introduce unimodal graphs where features serve as nodes, and edges are established by computing cosine similarity between nodes. A pictorial representation of graph construction and feature reconstruction method is shown in \autoref{fig:graph}. Subsequent sections elaborate on graph construction and feature refinement.
\subsubsection{Graph Construction}
As shown in Algorithm \ref{alg-graph}, the graph is constructed by calculating cosine similarity between nodes and forming an adjacency matrix.

Let us assume \({\mathcal{M}^x}=\{E_i^x\}_{i=1}^R\), where \(\mathcal{M}^x \in \mathbb{R}^{R \times D}\) is the embedding matrix, \(x \in \{\text{txt}, \text{img}\}\) denotes one of the modaities, $R$ denotes the total number of embeddings, and $D=512$ represents the dimension. We first normalize $\mathcal{M}^x$ with L2 normalization, ${\mathcal{M}_\text{norm}^x} = \{\mathcal{E}_i^x\}_{i=1}^R$.

We compute cosine similarity between each pair of embeddings to obtain the similarity matrix, $\text{sim}^x$ as shown in \autoref{eq:cosine}:

\begin{equation}
\label{eq:cosine}
    \text{sim}_{ij}^x = \text{cos\_sim}({{\mathcal{E}}}_i^x, {{\mathcal{E}}}_j^x) = \frac{{{{\mathcal{E}}}_i^x \cdot {{\mathcal{E}}}_j^x}}{{\| {{\mathcal{E}}}_i^x \|_2 \cdot \| {{\mathcal{E}}}_j^x \|_2}}
\end{equation}

where, $\text{sim}_{ij}^x$ is the similarity score of two embeddings. Upon computing similarities between every pair, we get a square matrix of size $R \times R$. 

\begin{algorithm}[!h]
  \caption{Graph Construction}
   \label{alg-graph}
  \begin{tabular}{ll}
    \textit{Input:} & $\mathcal{M}$: Embeddings matrix \\
                    & $thr$: Similarity threshold \\
    \textit{Output:} & $\text{Graph } \mathcal{G}$ \\
  \end{tabular}
  \begin{algorithmic}[1]
    \Function{cos\_sim}{$\tilde{\text{V}}_1$, $\tilde{\text{V}}_2$}
      \State \Return $\frac{\tilde{\text{V}}_1 \cdot \tilde{\text{V}}_2}{\|\tilde{\text{V}}_1\| \cdot \|\tilde{\text{V}}_2\|}$
    \EndFunction

    \Function{graph\_construction}{$\mathcal{M}$, $thr$}
      \State $\text{Adj} \leftarrow \text{[ ][ ]}$
      \For{$i \gets 1$ to $\text{len}(\mathcal{M})$}
        \For{$j \gets 1$ to $\text{len}(\mathcal{M})$}
          \State $\text{sim} \gets \text{cos\_sim}(\mathcal{M}_i, \mathcal{M}_j)$
          \If{$\text{sim} > \text{threshold}$}
            \State $\text{Adj}[i][j] \gets 1$
          \EndIf
        \EndFor
      \EndFor
      \State \Return $\text{Graph } \mathcal{G} 
  (\text{Adj})$
    \EndFunction
  \end{algorithmic}
\end{algorithm}

Next, we compute an adjacency matrix $\text{Adj}^x$ by putting an edge between two nodes based on the similarity threshold, $thr$. 
\begin{equation}
    \text{Adj}_{ij}^x =
    \begin{cases}
    1, & \text{if } \text{sim}_{ij}^x > {thr} \\
    0, & \text{otherwise}
    \end{cases}
\end{equation}

$\text{Adj}_{ij}^x = 1$ signifies the existence of an edge whereas $\text{Adj}_{ij}^x = 0$ signifies that there is no edge between any two arbitrary nodes i and j.


\subsubsection{Feature Reconstruction}
Unimodal features are reconstructed to enhance their representation. We employ GraphSAGE, an inductive learning-based graph neural network for feature reconstruction~\cite{hamilton2017inductive}. GraphSAGE can generalize its learning to unseen nodes during training (inductive). GraphSAGE acquires node embeddings by aggregating information from the neighborhood of each node. Different aggregation functions can be used in GraphSAGE, such as mean, max, sum, and min aggregation. We adopt mean aggregation as shown in \autoref{eq:mean_agg}:

\begin{equation}
\label{eq:mean_agg}
    \text{mean}(N_k, \mathcal{G}) = \frac{1}{|\mathcal{N}(N_k, \mathcal{G})|} \sum_{N_i \in \mathcal{N}(N_k, \mathcal{G})} \mathcal{E}_i
\end{equation}

where, $\mathcal{G}$ represents the graph, \(\mathcal{N}(N_k, \mathcal{G})\) is the set of neighbors for the node $N_k$, and $\mathcal{E}$ is the embedding such that $\mathcal{E} \in \mathcal{M}$. 
 Aggregating neighboring feature vectors, the current representation of node $\mathcal{E}_k$ corresponding to $N_k$ is subsequently combined through concatenation with the aggregated neighborhood vector. The concatenated features are passed through a fully connected layer utilizing learnable weight $W$, an activation function $\sigma$, and bias $b$. 

\begin{equation}
\label{eq:graph-sage-conv}
    \mathcal{H} = \sigma(\text{concat}(\text{mean}(N_k,\mathcal{G}), {\mathcal{E}_k}) \cdot W) + b
\end{equation}

Finally, we concatenate image and text features, $\mathcal{H}^{\text{img}}$ and $\mathcal{H}^{\text{txt}}$, to form a multimodal relation feature vector denoted as $\mathcal{\tilde{H}}^{\text{img-txt}} \in \mathbb{R}^{512}$, where, $\mathcal{H}^{\text{img}} \in \mathbb{R}^{256}$ and $\mathcal{H}^{\text{txt}} \in \mathbb{R}^{256}$ represent graph-reconstructed image and text features, respectively.



\subsection{\textit{Content-specific Feature Learning Module (CFLM)}}
In this module, all the content-specific features described in the
in \autoref{sec:cflm_feat} are learned. These features include image-based features and text-based features. Image-based features include  NSFW features $f^{\text{nsfw}} \in \mathbb{R}^{5}$ and caption features $f^{\text{cap}} \in \mathbb{R}^{512}$, whereas text-based features include toxicity features $f^{\text{tox}} \in \mathbb{R}^{6}$ and MSL score $f^{\text{msl}} \in \mathbb{R}^{1}$. These features are first concatenated to get a joint representation $\tilde{f} \in \mathbb{R}^{524}$.
Next, the vector $\tilde{f}$ is passed through a fully connected layer to get the learned joint vector.

\begin{equation}
    \tilde{\mathcal{F}} = \sigma(W_f \cdot \tilde{f} + b_f) 
\end{equation}
Here, $\tilde{\mathcal{F}} \in \mathbb{R}^{256}$ is learned joint vector for content-specific features, \(W_f \in \mathbb{R}^{256 \times 512}\) is weight matrix, $\sigma$ denotes activation function and $b_f$ is bias. Joint vector $\tilde{\mathcal{F}}$ is then concatenated with output vectors of MANM and GFRM modules for joint learning and classification.

\subsection{\textit{Classification Module}}
The classification module generates final predictions of given multimodal content. During testing, it applies test-time augmentation, while training is performed only on original samples.
The classification module first concatenates the features obtained through MANM, GFRM, and CFLM as shown in \autoref{eq:joint_all}: 
\begin{equation}
\label{eq:joint_all}
    \tilde{\mathcal{Y}}^{\text{joint}} = \text{concat}(\mathcal{\tilde{X}}^{\text{img-txt}}, \mathcal{\tilde{H}}^{\text{img-txt}}, \tilde{\mathcal{F}})
\end{equation}
where, $\tilde{\mathcal{Y}}^{\text{joint}} \in \mathbb{R}^{1280}$, represents the final joint multimodal vector.
The integrated representation of these features is passed through a series of three fully connected layers, which have output dimensions of
1024, 512, and 256 respectively. Each fully connected layer is followed by a dropout layer of 0.3. A sigmoid-activated fully connected layer is used to generate the final probability. During training, we use Adam optimizer, a batch size of 128, and a learning rate of 1e-4. During testing, we employ TTA which is discussed in the next subsection.
\subsubsection{\textit{Test-Time Augmentation (TTA)}}
\begin{figure}[htbp]
    \centering
    \includegraphics[width=8.5cm, height=2.98cm]{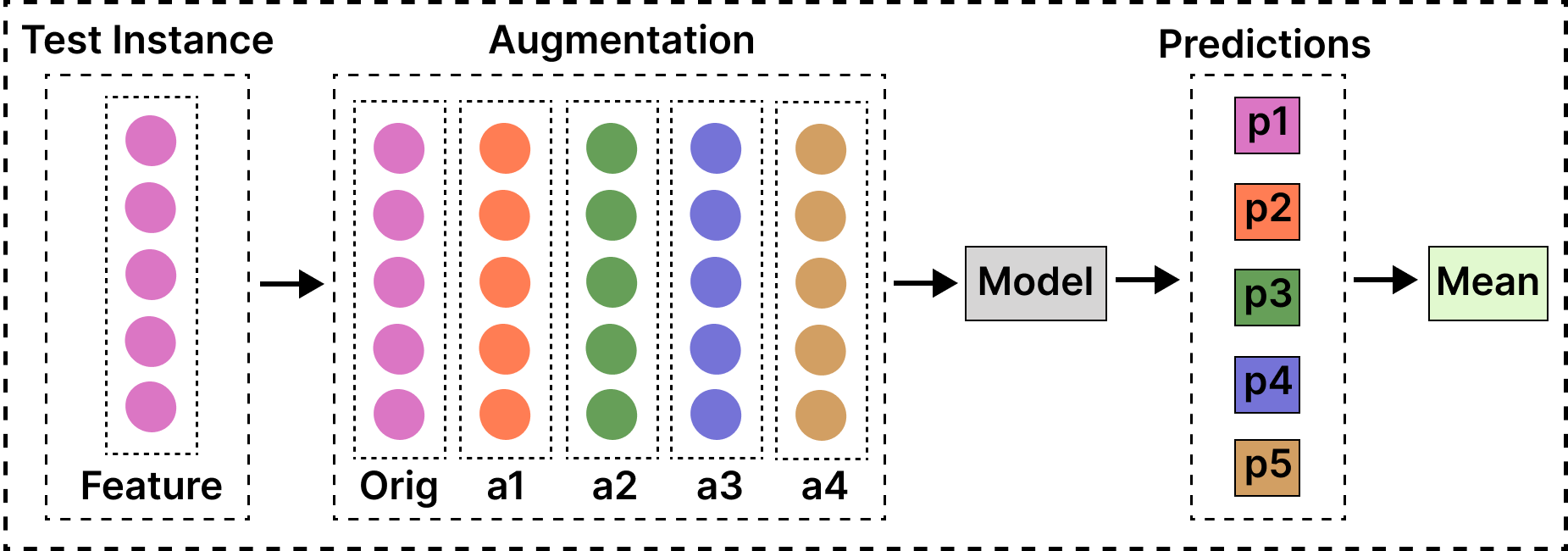}
    \caption{Test-Time Augmentation}
    \label{fig:TTA}
\end{figure}

Test-Time augmentation (TTA) is performed during testing by augmenting multiple versions of a test sample and the aggregating of the predictions on these samples to get the final prediction. TTA has been explored in Computer Vision \cite{shanmugam2021better, oza2023breast} and text classification tasks \cite{cohen2023enhancing,shleifer2019low}. \autoref{fig:TTA} illustrates augmentation and final prediction generation approach at test-time.

We employ feature space data augmentation using additive random perturbations \cite{kurata2016labeled}. Algorithm-\ref{alg-TTA} represents the steps for the additive random perturbation method. To apply additive random perturbations to feature vectors, we generate a vector with random values between $-p$ and $+p$. Subsequently, an element-wise sum is performed between the feature vector and random vector that provides a new synthetic feature vector. The same approach is used to generate synthetic feature vectors for both modalities. To make sure that synthetic vectors are not very similar or very dissimilar to the original feature vector, we take only those synthetic vectors that have cosine similarity between 0.6 to 0.7 including the boundary conditions. For cosine-similarity computation between two 2D vectors, we take a mean of 2D vectors to get 1D sequence vectors.
 
\begin{algorithm}[!h]
    \caption{Additive Random Perturbation}
    \label{alg-TTA}
    \begin{tabular}{ll}
    \textit{Input:} & $H^{\text{txt}}$:  Text features of test set\\
    &  $H^{\text{img}}$:  Image features of test set \\
  \textit{Output:} 
  & $S^{\text{txt}}_{\text{aug}}, S^{\text{img}}_{\text{aug}}$: Augmented test sets\\
  \textit{function:} & TTA($H^{\text{txt}}$,$H^{\text{img}}$)
    \end{tabular}
    \label{algo:data_aug}
    \begin{algorithmic}[1]
    \State{$ 
S^{\text{txt}}_{\text{aug}}, S^{\text{img}}_{\text{aug}} \gets $\{ \}}

\State{\textbf{for each 
$h^{\text{txt}} \in H^{\text{txt}}$ and $h^{\text{img}} \in H^{\text{img}}$   do}
}
\State{\hspace{3mm}\textbf{for  
$i > n$ do} 
} 
\State{\hspace{6mm}$ 
\mathcal{R}_t \gets rand\_vector(\text{len}(h^{\text{txt}}),p)$}
\State{\hspace{6mm}$
\mathcal{R}_v \gets rand\_vector(\text{len}(h^{\text{img}}),p)$
}
\State{\hspace{6mm}$
S^{\text{txt}}_{\text{aug}} \gets h^{\text{txt}} + \mathcal{R}_t$
}
\State{\hspace{6mm}$
S^{\text{img}}_{\text{aug}} \gets h^{\text{img}} + \mathcal{R}_v$
}
\State{\hspace{3mm}\textbf{end for}}
\State{\textbf{end for}}
    \State{\Return $S^{\text{txt}}_{\text{aug}},S^{\text{img}}_{\text{aug}}$}
    \end{algorithmic}
    \end{algorithm}

We generate four additional synthetic feature embeddings for each test sample in order to apply TTA. During test-time augmentation, we are able to generate the synthetic embeddings; however, for content-specific features, we use the same values as extracted for the original sample. 

\section{Experimental Evaluations}
The experimental setup is initially explained in this section, followed by the experimental results to illustrate the effectiveness of the proposed method.

\subsection {Experimental Setup}
In this section, we provide an overview of the datasets utilized for experimental evaluations. Subsequently, we delve into the discussion of  evaluation metrics and distinct methods employed for comparative analysis.

\subsubsection{\textit{Summarization of Datasets}}\label{label:dataset_summary}
We perform misogyny detection and sexism detection tasks on two different datasets, namely, Multimedia Automatic Misogyny Identification (MAMI) and MMHS150K. The misogyny detection task is performed on the MAMI dataset that is provided in Task-5 of SemEval-2022 competition~\cite{fersini2022semeval}. For sexism detection, we utilize the MMHS150K dataset~\cite{gomez2020exploring}.
\begin{enumerate}[ label=\alph*) ]
\item{
        \textit{MAMI}:
         \hspace{2mm}The MAMI dataset has 10,000 annotated memes. The dataset has 5,000 misogynous and 5,000 non-misogynous memes in the training set. The testing set has a total of 1,000 samples that are equally distributed in the misogynous and non-misogynous classes.}

\item{
        \textit{MMHS150K}:
        \hspace{2mm}The MMHS150K dataset comprises 1,50,000 tweets with text and images collected from Twitter. It focuses on tweets containing 51 Hatebase terms associated with hate speech. Among different hate speech classes, we utilize samples only from the sexism class, which has 3,494 samples. For the non-hate class, we take a total of 10,000 samples. The total samples, 13,494, are then split into a ratio of 80:20 for training and testing sets.}
\end{enumerate}

\subsection {Evaluation Metrics}
We evaluate the efficacy of the proposed approach and conduct a comparative analysis based on metrics such as Accuracy, Precision, Recall, and F1 Score.  Let us assume TP denotes True Positive , TN denotes True Negative, FP is False Positive, and FN is False Negative, then:

\begin{equation}
\text{Accuracy (Acc)}  = \frac{TP + TN}{TP + TN + FP + FN}
\end{equation}
\vspace{-5mm}
\begin{equation}
\text{Precision (P)} = \frac{TP}{TP + FP}
\end{equation}
\vspace{-5mm}
\begin{equation}
\text{Recall (R)} = \frac{TP}{TP + FN}
\end{equation}
\vspace{-5mm}
\begin{equation}
\text{F1 Score (F1)} = \frac{2 \times \text{Precision} \times \text{Recall}}{\text{Precision} + \text{Recall}}
\end{equation}

We use macro-averaged F1, which is computed by taking the arithmetic mean of F1-scores of classes.

\subsubsection{\textit{Comparison Methods}}
    We conduct a comparative analysis with the following approaches for misogyny detection:
\begin{enumerate}[ label=\alph*) ]
\item{\textit{Unimodal Methods:}
We employ a state-of-the-art transformer model, specifically BERT for textual data and MASK RCNN RESNET-101 for images, to independently generate embeddings or feature representations for each modality.  BERT is applied to capture contextual information within the text, while MASK-RCNN RESNET-101 excels in extracting meaningful region-based features from images.}

\item {\textit{Multimodal Methods:}
We use a combination of images and text for misogyny detection. We compare the performance of the proposed method on both datasets with multimodal methods that include the Late Fusion method with BERT and MASK RCNN RESNET-101 features, VisualBERT, CLIP, and ViLT. We also compare the performance of the proposed approach on MAMI with existing research works such as BERT+ViT, RIT Boston, DD-TIG, and SRCB. For comparison on the MMHS150K dataset, we use BERT+ViT, FCM, MSKAV+KDAC, and ARC-NLP. }

\end{enumerate}

\begin{table*}[htbp]
\centering
\caption{Performance comparison over MAMI dataset\label{tab:mami_results}}
\begin{tabular*}{\linewidth}{@{\extracolsep{\fill}} clcccccc }
\toprule
 & Method & Acc &P &R & F1 \\
\midrule
\multirow{2}{*}{Unimodal } 
 & BERT \cite{devlin2018bert}& 68.16 & 67.84 & 69.20 &68.16\\
 & Mask R-CNN \cite{he2017mask} & 63.16 & 65.13 & 56.80 &63.01 \\
\midrule
\multirow{8}{*}{Multimodal} 
 & Late Fusion (MASK RCNN + BERT)~\cite{gandhi2023multimodal} & 69.26 & 68.03 & 72.76 &69.23 \\
 & VisualBert~\cite{li2019visualbert} & 73.85 & 69.48 & 78.80 &73.94\\
 & CLIP~\cite{radford2021learning} & 78.10 & 76.77 & 80.06 &78.08 \\
 & ViLT~\cite{kim2021vilt} & 73.20 & 66.66 & 92.80 &73.12\\
& BERT + ViT~\cite{singh2023female} &86.20 & - &88.10 &87.40 \\
& RIT Boston~\cite{chen2022rit}  &- &- &- &77.80 \\
& DD-TIG~\cite{zhou2022dd}  &79.40 &- &- &79.30 \\
& SRCB~\cite{zhang2022srcb}  &- &- &- &83.40 \\
\midrule
\multirow{1}{*}{} 
 & Proposed & \textbf{87.42} & \textbf{83.81} & \textbf{89.90} &\textbf{87.95}\\
\bottomrule
\end{tabular*}
\end{table*}

\subsection{Results}
This section presents experimental results and comparisons. 
First, a comparison of the proposed approach with different unimodal and multimodal methods is presented. Subsequently, an ablation analysis is conducted, exploring the influence of various components and features.

\subsubsection{	Performance Comparison over MAMI Dataset}
In this section, we undertake a thorough comparison between our proposed method and existing approaches on the MAMI dataset. This comparative analysis seeks to highlight the distinctive strengths and advantages of our methodology in relation to the prevailing unimodal and multimodal techniques. The results are shown in \autoref{tab:mami_results}.

The proposed model outperforms the unimodal approaches as it learns from both texts as well images. These features allow our proposed model to achieve an improvement of 19.79\% in terms of macro-F1 over the unimodal BERT approach and 24.94\% in terms of macro-F1 over the unimodal MASK RCNN RESNET-101 approach. In comparison to multimodal approaches, the proposed method shows significant performance gains on different evaluation metrics.
 The proposed model improves the macro-F1 score by 18.72\%, 14.01\%, 9.87\%, and 14.83\% for the late-fusion method, VisualBERT, CLIP, and VILT respectively.
Our proposed model outperforms the existing state-of-the-art works. In terms of macro-F1, it achieves an absolute improvement of  4.55\% over the SRCB,  8.55\% over the DD-TIG, 0.55\% over the BERT +ViT
and 10.15\% over the RIT Boston. The comparison results on the MAMI dataset denote the efficacy of the proposed framework in misogyny detection. 
The performance gain is due to the fact that the proposed employs a dedicated graph-based module for unimodal feature refinement and a dedicated attention module for cross-modal features. It also utilizes the synergy of other features such as toxicity features, misogyny lexicons, NSFW model features, and caption embeddings. Additionally, to generalize the predictions, we employ TTA.

\subsubsection{Performance Comparison over MMHS150K Dataset} The MMHS dataset contains multimodal content that has two classes: sexism and non-sexism. The proposed approach outperforms the existing methods by a significant margin in terms of accuracy, precision, recall, and F1 score. The results are shown in \autoref{tab:mmhs_results}. The proposed model outperforms the unimodal approaches by achieving an improvement of 17.76\% and 16.39\% in terms of macro-F1 score over the unimodal BERT and the unimodal MASK RCNN RESNET-101 approach, respectively.  The results outline the importance of using both modalities. The proposed model outperforms the multimodal approaches as it learns from content-specific features, test time augmentation, MCAN, and graph-based features. This allows it a deeper understanding of misogyny-related content. The proposed model improves macro-F1 score by 13.15\%, 11.66\%, 2.71\%, and 12.78\% for late-fusion, VisualBERT, CLIP, and VILT, respectively.

Our proposed model outperforms the existing state-of-the-art methods. It achieves an absolute improvement of  7.19\%, 9.93\%, 15.27\%, and 12.20\% in terms of accuracy, precision, recall, and macro-F1 score over BERT +ViT. It also gains an improvement of  6.82\%, 10.79\%, 15.19\%, and 12.99\% in terms of accuracy, precision, recall and macro-F1 score over the FCM, 3.89\%, 1.94\%, 4.19\% and 3.17\% in terms of accuracy, precision, recall and macro-F1 over MSKAV+KDAC and 0.72\%, 1.80\%, 2.28\% and 2.36\% over the ARC-NLP. Overall, the proposed approach outperforms the existing method across multiple evaluation metrics.

\begin{table*}[htbp]
\centering
\caption{Performance comparison over MMHS150K dataset\label{tab:mmhs_results}}
\begin{tabular*}{\linewidth}{@{\extracolsep{\fill}} clcccccc }
\toprule
 & Method & Acc &P &R & F1 \\
\midrule
\multirow{2}{*}{Unimodal } 
 & BERT \cite{devlin2018bert}& 78.84 & 73.97 & 67.01 &68.86\\
 & Mask R-CNN \cite{he2017mask} &76.84 &70.37 &70.09 &70.23 \\
\midrule
\multirow{8}{*}{Multimodal} 
 & Late Fusion (MASK RCNN + BERT)~\cite{gandhi2023multimodal} &80.69 &75.88 & 72.02 &73.47 \\
 & VisualBert~\cite{li2019visualbert} & 81.58 & 77.26 & 73.85 &74.96\\
 & CLIP~\cite{radford2021learning} & 86.63 & 82.58 & 85.83 &83.91 \\
 & ViLT~\cite{kim2021vilt} & 81.13 & 76.56 & 72.62 &73.84\\
& BERT + ViT~\cite{singh2023female} &80.75 &75.33 &73.18 &74.42 \\
& FCM~\cite{gomez2020exploring}  &81.12 &74.47 &73.26 &73.63 \\
& MSKAV+KDAC~\cite{CHHABRA2023106991}  &84.05 &83.32 &84.26 &83.45 \\
& ARC-NLP~\cite{thapa2023multimodal}  &87.22 &83.46 &86.17 &84.26 \\
\midrule
\multirow{1}{*}{} 
 & Proposed & \textbf{87.94} & \textbf{85.26} & \textbf{88.45} &\textbf{86.62}\\
\bottomrule
\end{tabular*}
\end{table*}

\subsubsection{Ablation Analysis}
In the ablation analysis, we analyze the impact of important components and features on misogyny detection over the MAMI dataset.

a) \textit{\textbf{Influence of Different Modules}}: To evaluate the impact of attention and graph-based modules, we first remove one of these components while keeping the other components intact and then we evaluate its performance on all four evaluation metrics. We then compare the results with the originally proposed model. The results are presented in \autoref{tab:abl-module}. 
\begin{table}[HBP]
    \centering
    \caption{Ablation analysis on different modules \label{tab:abl-module}}
    \begin{tabular}{lccccccccc}
    \toprule
     & Acc  & P & R & F1  \\
    \midrule
    w/o MANM & 83.25  & 81.56 & 85.41 & 83.29 \\
    w/o GFRM & 84.40  & 82.11 & 87.25 & 84.48 \\
    w/o CFLM & 84.91  & 82.26 & 87.38 & 84.73 \\
    Proposed & 87.42 & 83.81 & 89.90 & 87.95 \\
    \bottomrule
    \end{tabular}
\end{table}
Without the attention-based module MANM, accuracy, precision, recall, and macro-F1 reduce
by 4.17\%, 2.25\%, 4.49\%, and 4.66\%, respectively. The graph-based module GFRM also has a significant impact on the model's performance. Without graph-based module accuracy, precision, recall, and macro-F1 score decrease by 3.02\%, 1.70\%, 2.65\%, and 3.47\%, respectively. Removal of all the content-specific features reduces the model's performance by 2.51\%, 1.55\%, 2.52\%, and 3.22\% in terms of accuracy, precision, recall, and macro-F1.

\begin{figure}[htbp]
    \centering
    \includegraphics[width=8.5cm, height=5.18cm]{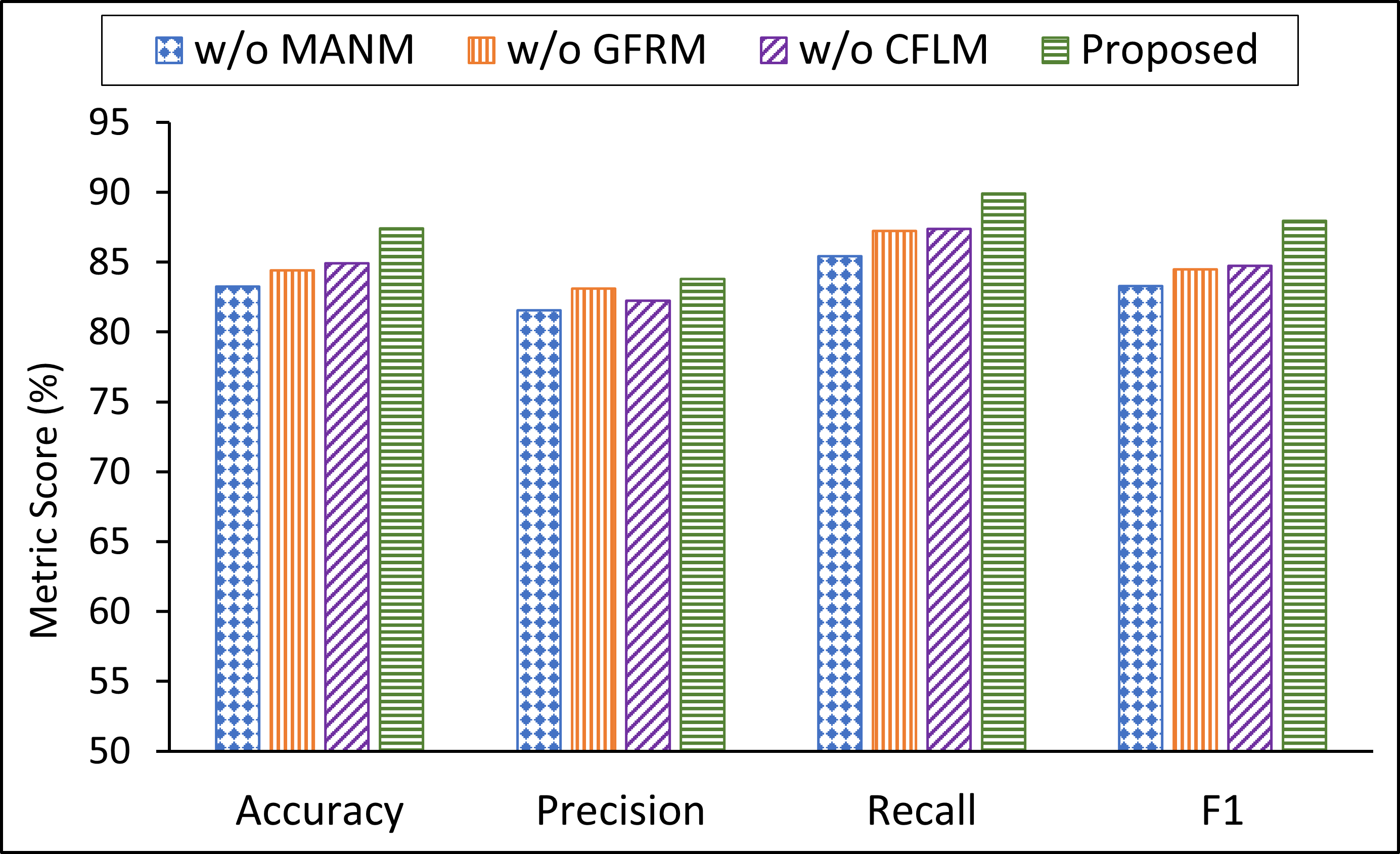}
    \caption{Influence of different modules}
    \label{fig:abla_module}
\end{figure}

\autoref{fig:abla_module} shows a pictorial representation of module influence results. Using attention allows our model to capture context-aware relationships between images and text, enabling it to understand dependencies between relevant regions and words. This empowers the model to concentrate on particular segments of the input by assigning weights determined by their pertinence. The graph-based module is able to provide modality-specific refined features, where content-specific features provide useful insights into the given multimodal that help train the model.\\

\begin{figure}[htbp]
    \centering
    \includegraphics[width=8.5cm, height=5.18cm]{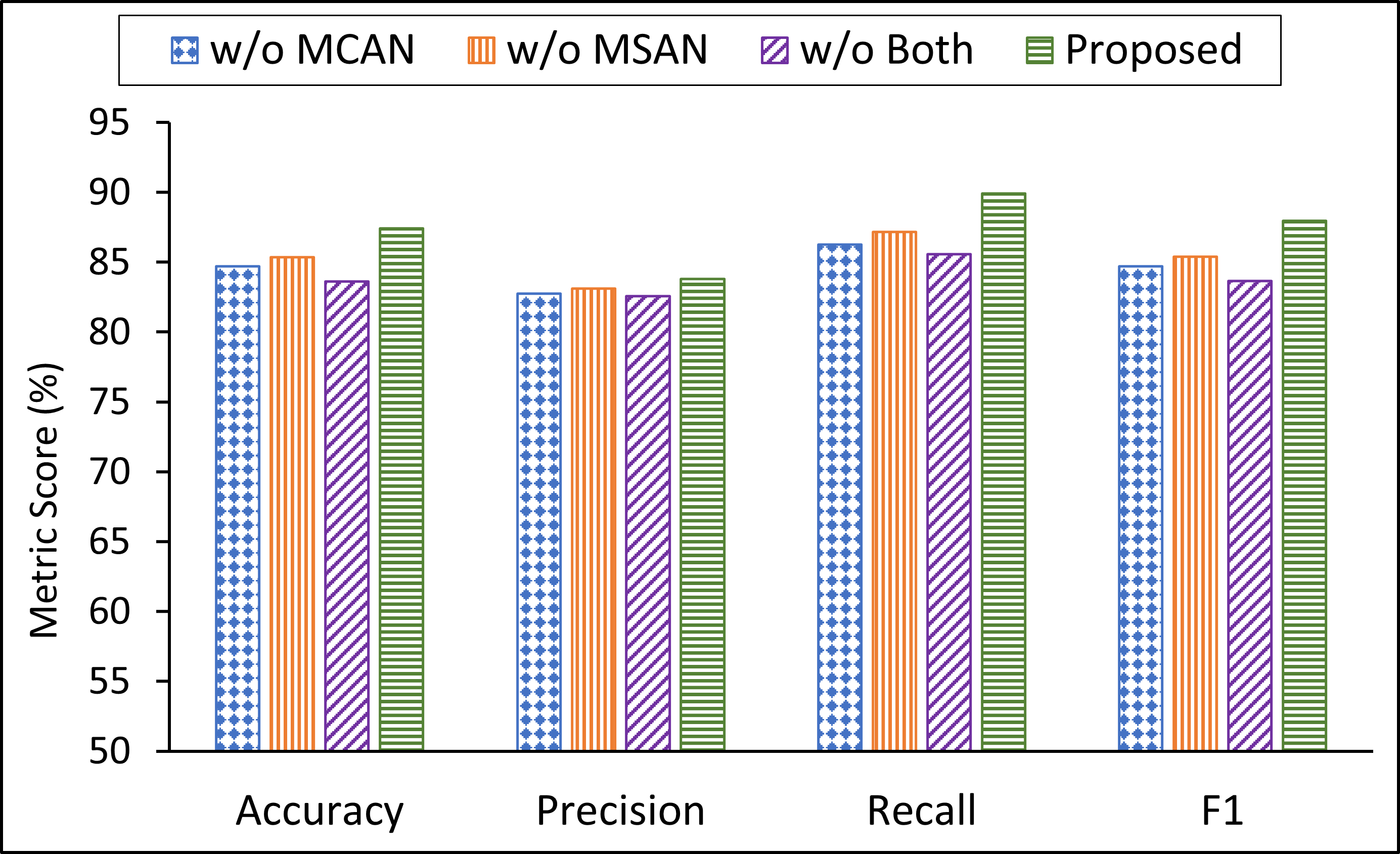}
    \caption{Influence of different attention mechanisms}
    \label{fig:abla_attention}
\end{figure}
b) \textit{\textbf{Influence of Different Attention Mechanisms}}: The proposed approach utilizes two attention mechanisms, MSAN and MCAN. We analyze the influence of MSAN and MCAN for misogynous content identification. To evaluate the performance of an attention mechanism, we remove that attention and analyze the reduction in performance. Next, we evaluate the combined influence of both attention mechanisms by removing both. When we remove both attention mechanisms, we still use BERT and MASK RCNN features using max-pooling. We evaluate the performance in terms of precision, recall, macro-F1, and accuracy. The results are shown in \autoref{tab:abl-attention}. Without the self-attention accuracy, precision, recall, and macro-F1 reduce by 2.08\%, 0.69\%, 2.74\% and 2.57\% respectively. Whereas without MCAN accuracy, precision, recall, and macro-F1 score decrease by 2.70\%, 1.07\%, 3.64\%, and 3.24\% respectively. When both attention mechanisms are removed, we observe a significant decrease in accuracy, precision, recall, and macro-F1, i.e., 3.79\%, 1.23\%, 4.34\%, and 4.30\%, respectively. The results demonstrate that MSA and MCAN attentions allow the model to dynamically adjust its focus based on the surrounding context, enabling more informed and contextually relevant processing of information. \autoref{fig:abla_attention} presents a graphical visualization of the attention-based ablation results.

\begin{table}[H]
    \centering
    \caption{Ablation analysis on attention mechanisms} \label{tab:abl-attention}
    \begin{tabular}{lccccccccc}
    \toprule
     & Acc  & P & R & F1  \\
    \midrule
    w/o MCAN & 84.72  & 82.74 & 86.26 & 84.71 \\
    w/o MSA & 85.34  & 83.12 & 87.16 & 85.38 \\
    w/o Both & 83.63  & 82.58 & 85.56 & 83.65 \\
    Proposed & 87.42 & 83.81 & 89.90 & 87.95 \\
    \bottomrule
    \end{tabular}
\end{table}

c) \textit{\textbf{Influence of Modalities}}: The influence of modalities on the model's performance refers to using either only text or only images for misogyny detection while keeping other components the same as before. The results are shown in \autoref{tab:abl-modality}. For text-only and image-only methods, the MANM module can not be used because it requires cross-attention. Content-specific features are used only for the modality that we are comparing.

Removal of MANM, embeddings of other modality, and content-specific features of other modality affect the performance. While using only text modality, the accuracy reduces by 14.94\%, precision reduces by 18.76\%, recall decreases by 5.70\% and macro-F1 score decreases by 15.10\%. These results demonstrate that images help significantly in misogyny detection.
Next, we measure the performance using only images. While using only images, the accuracy reduces by 18.32\%, precision reduces by 19.39\%, recall reduces by 4.39\% and macro-F1 score decreases by 18.72\%.

\begin{figure}[htbp]
    \centering
    \includegraphics[width=8.5cm, height=5.18cm]{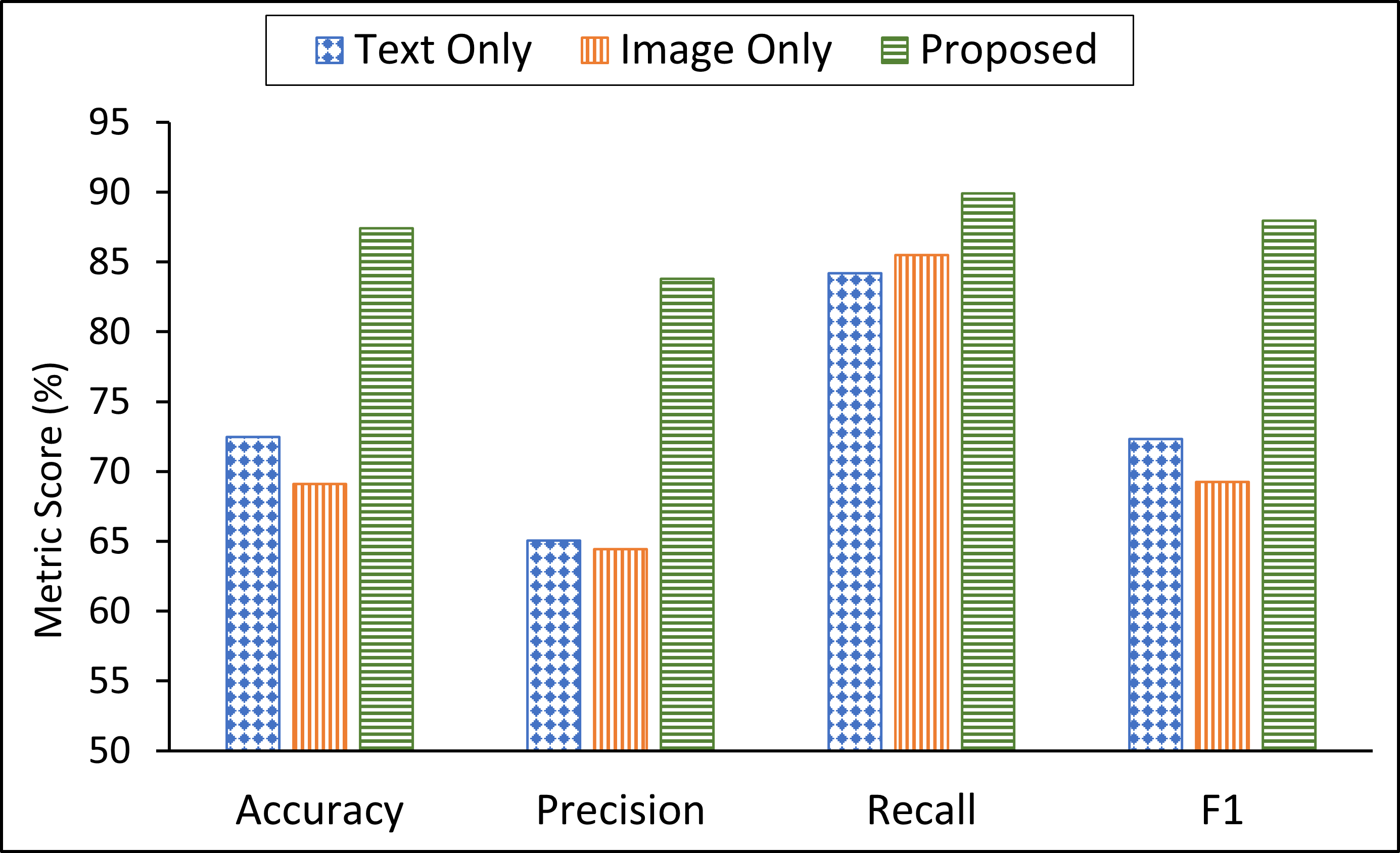}
    \caption{Influence of different modalities}
    \label{fig:abla_modality}
\end{figure}

As shown in \autoref{fig:abla_modality}, the proposed model outperforms the text-only and image-only approaches as it learns from both texts as well as images for classification.

\begin{table}[htbp]
    \centering
    \caption{Ablation analysis on modalities \label{tab:abl-modality}}
    \begin{tabular}{lccccccccc}
    \toprule
     & Acc  & P & R & F1  \\
    \midrule
    Text only & 72.48  & 65.05 & 84.20 & 72.31 \\
    Image only & 69.10  & 64.42 & 85.51 & 69.23 \\
    Proposed & 87.42 & 83.81 & 89.90 & 87.95 \\
    \bottomrule
    \end{tabular}
    \end{table}
    
d) \textit{\textbf{Influence of Additional Features}}: We examine the influence of supplementary features on the performance of our method. In order to assess the impact of a feature, we remove that feature and get the results. These supplementary features include NSFW features, toxicity features, caption embeddings, and MSL score. Additionally, we also show the influence of TTA. The results of this ablation are shown in \autoref{tab:abl-feature}.
\begin{table}[h]
    \centering
    \caption{Ablation analysis on additional features \label{tab:abl-feature}}
    \begin{tabular}{lccccccccc}
    \toprule
     & Acc  & P & R & F1  \\
    \midrule
    w/o NSFW  & 86.62 & 83.56 & 89.15 & 86.98 \\
    w/o Toxicity & 86.23 & 83.45 & 88.58 & 86.24 \\
    w/o Caption & 85.52 & 83.12 & 87.90 & 85.61 \\
    w/o MSL Score & 86.43 & 83.70 & 88.56 & 86.81 \\
    w/o TTA & 85.38 & 82.27 & 87.48 & 85.35 \\
    Proposed & 87.42 & 83.81 & 89.90 & 87.95 \\
    \bottomrule
    \end{tabular}
\end{table}

\begin{figure}[htbp]
    \centering
    \includegraphics[width=8.5cm, height=5.18cm]{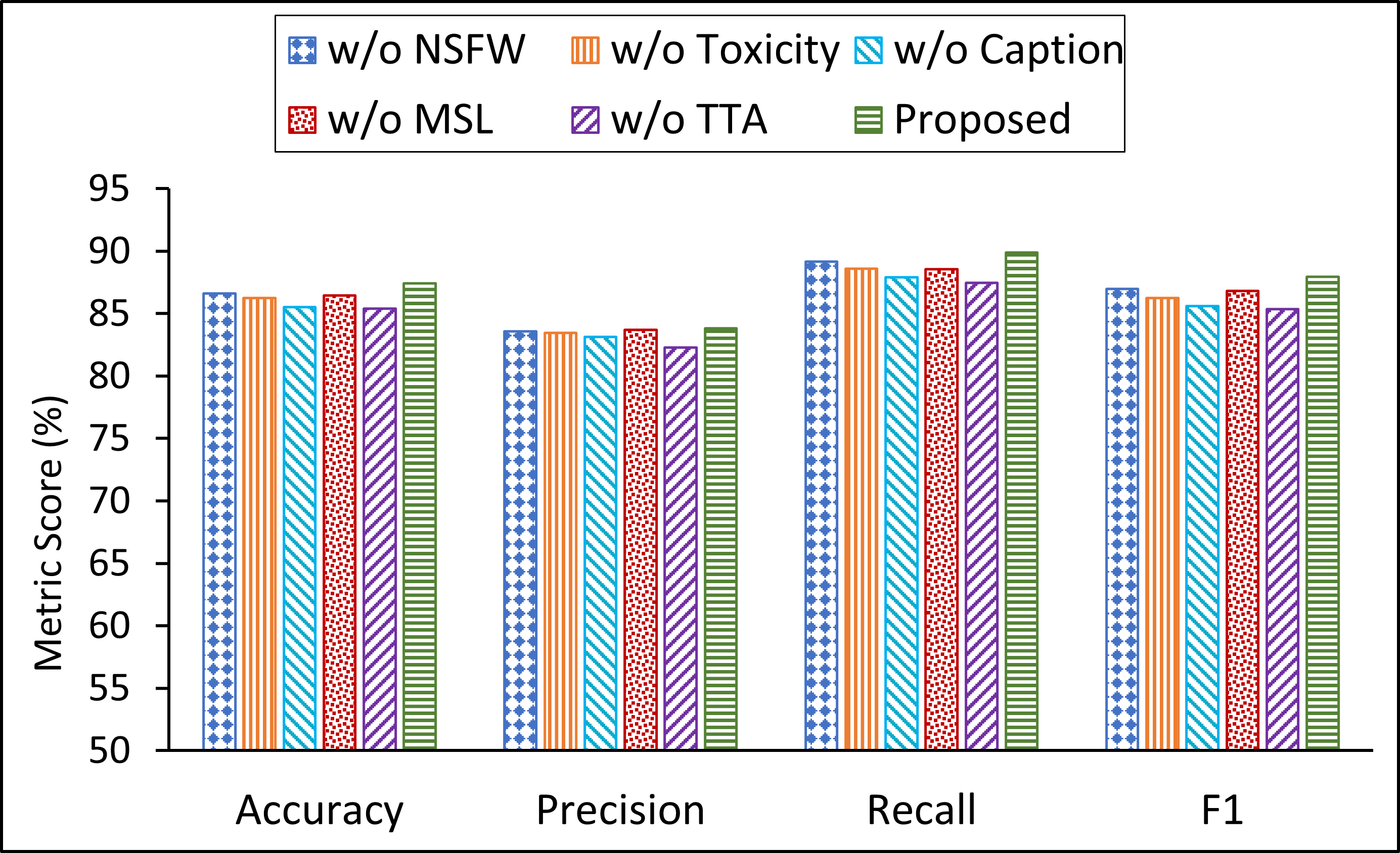}
    \caption{Influence of additional features}
    \label{fig:abla_features}
\end{figure}
The improvements in terms of accuracy, precision, recall, and macro-F1 score by using these features are as follows: 0.80\%, 0.25\%, 0.75\%, 0.97\% (NSFW  features), 1.19\%, 0.36\%, 1.32\%, and 1.71\% (toxicity features),
1.9\%, 0.69\%, 2.00\%, and 2.34\% (caption embeddings),
1.00\%, 0.11\%, 1.34\%, and 1.14\% (MSL score), and 2.04\%, 1.54\%, 2.42\%, and 2.60\% (TTA). A graphical representation of these results is given \autoref{fig:abla_features}. 
These features allow the proposed model to understand the context and useful insights of the content by utilizing both modalities, hence improving the overall performance of the model.

\begin{figure*}
    \centering
    \includegraphics[width=16 cm, height=8.85cm]{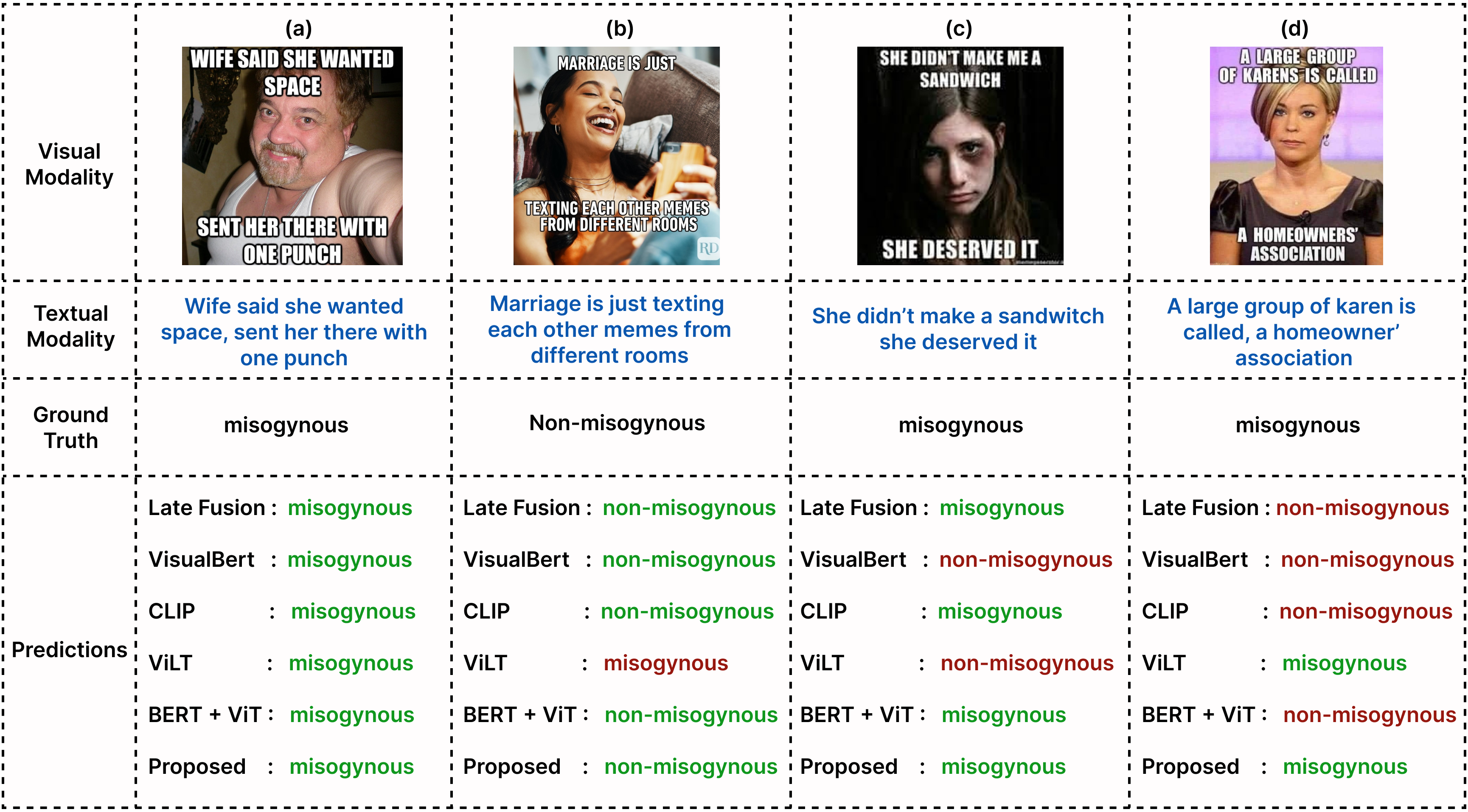}
    \caption{Qualitative analysis is based on 4 different memes. The first two memes are relatively simpler to predict than the last two, as the multimodal methods struggled to predict the last two correctly. }
    \label{fig:qual}
\end{figure*}

\subsection{Qualitative Analysis}
This section presents a qualitative analysis of the proposed framework for different memes. Analysis is based on its comparison with predictions of other multimodal methods by examining various aspects of the model's performance in relation to its counterparts. A pictorial representation of qulatitaive analysis in given in \autoref{fig:qual}.

Figure 10 (a), presents a multimodal content which has strong language against women. The content reads "wife said she wanted space, sent her there with one punch" with a smiling man's image. The overall content depicts violence against women hence it is labeled as misogynous. Because of its strong language, all the methods predict it correctly.

In Figure 10 (b), image portrays a laughing lady which conveys a positive emotions. The text in the image reads "marriage is just texting each other memes from different rooms" which contains a humour and irony. Hence the memes presents a humor without any offensive element against women. ViLT for some reason predict it as misogynous, this is due to the fact that in a few training samples the word marriage is associated with misogyny class, therefore ViLT is not able to detect the overall positive sentiment of the image. Most of the methods, including the proposed one, predict it correctly as non-misogynous.

Figure 10 (c) shows a meme that depicts violence against women. The visual content shows a woman with bruised face giving a very negative sentiment and the text conveys that she is hit due to something she does not agree to do. VisualBert and ViLT do not predict it correctly, while the others are able to capture the relationship between the text and image.

The content presented in the Figure 10 (d) reads "a large group of karens is called, a homeowner’ association" with visual content showing a lady with neutral expressions. Although the visual content does not depict any offense but in the text, the word karen is used as an offensive term against women. While most of the methods misclassify the content, the proposed method is able to predict it correctly as misogynous due to the incorporation of MSL score that is computed based on slangs and misogynous word. 

In conclusion, the qualitative analysis underscores the effectiveness of the proposed method in correctly predicting the presence of misogyny in memes. Through rigorous comparison with other model predictions, it becomes evident that the proposed framework demonstrates its strength by consistently and accurately identifying misogynistic content across diverse content.

\section{Implications}
The implications of an approach designed for misogyny detection are multifaceted and can have significant technical,  societal, and cultural ramifications. Here are some potential implications:

\subsection{Mitigation of Harmful Content}Our proposed misogyny detection approach can help identify and flag content that perpetuates gender-based discrimination, harassment, or violence against women. By identifying such content, platforms, and policymakers can take appropriate measures to mitigate its impact and create safer online environments.

\subsection{Psychological and Societal Impact} Exposure to misogynistic content can have detrimental effects on individuals' mental health and perpetuate harmful gender norms in society. By detecting and removing such content, our method may contribute to reducing the psychological harm experienced by women.

\subsection{Enhaced Multimodal Content Analysis}Our proposed approach can be helpful in other multimodal analyses where image and text play a crucial role, such as visual question answering. The porposed context-aware cross-attention is able capture intricate relation between the images and text.
\section{Conclusion}
\label{sec:sa_cnfw}
The proliferation of misogynistic and sexist content on social media has emerged as a noteworthy concern in recent times. In this article, we propose a novel multimodal framework for the detection of misogynistic and sexist content. Our proposed methodology involves feature learning via specialized modules, a novel context-aware attention module, a graph-based module for unimodal feature refinement, and a content-specific feature module. The final classification is achieved by integrating these features through a process of feature fusion, resulting in a unified representation. This joint feature vector is then processed through the classification network. Our findings indicate that context-aware attention helps get pertinent region-word features, resulting in a deeper understanding of the content. The graph-based method is able to comprehend and refine unimodal features. Moreover, we prepare a set of misogynistic lexicons to calculate the MSL score from the text that provides an initial insight into the content. During testing, we apply Test-Time Augmentation with additive random perturbations, contributing to enhanced generalization and more robust predictions across diverse inputs. We have performed extensive experiments on two multimodal datasets, MAMI and MMHS150K. Results demonstrate our proposed method achieves competitive results by outperforming existing state-of-the-art methods. The proposed approach could be used in filtering misogynistic and sexist content.

\printcredits \\ \\





\balance
\bibliographystyle{unsrtnat}

\bibliography{cas-refs}
\balance





\end{document}